\documentclass[letterpaper]{article} % DO NOT CHANGE THIS
\usepackage{aaai2027}  % DO NOT CHANGE THIS
\usepackage[hyphens]{url}  % DO NOT CHANGE THIS
\usepackage{graphicx} % DO NOT CHANGE THIS
\usepackage{natbib}  % DO NOT CHANGE THIS AND DO NOT ADD ANY OPTIONS TO IT
\usepackage{caption} % DO NOT CHANGE THIS AND DO NOT ADD ANY OPTIONS TO IT
\usepackage{algorithm}
\usepackage{algorithmic}

\usepackage{newfloat}
\usepackage{listings}
\DeclareCaptionStyle{ruled}{labelfont=normalfont,labelsep=colon,strut=off} % DO NOT CHANGE THIS
\floatstyle{ruled}
\newfloat{listing}{tb}{lst}{}
\floatname{listing}{Listing}

\usepackage{amsmath, amssymb}
\usepackage{mathtools}
\usepackage{multirow}
\usepackage{booktabs}
\usepackage{subcaption}
\usepackage{xcolor}

\newcommand{\defect}{\texttt{defect}}
\newcommand{\bydesign}{\texttt{by-design}}
\newcommand{\real}{\texttt{real}}
\newcommand{\intended}{\texttt{intended}}
\newcommand{\code}[1]{\texttt{\small #1}}

\newcommand{\pos}[1]{\left[#1\right]_{+}}

\title{From Handcrafted to Generative: RLVR-Optimized Icon Generation for Alexa+}
\author{
    Rohan Sequeira\textsuperscript{\rm 1}\thanks{Work done during an internship at Amazon.}, % Work done at internship
    Matthew Rosenberg\textsuperscript{\rm 2},
    Amin Mahmoudabadbozchelou\textsuperscript{\rm 2},\\
    Oleg Poliannikov\textsuperscript{\rm 2},
    Jonathan Rubin\textsuperscript{\rm 2},
}
\affiliations{

    \textsuperscript{\rm 1} University of Southern California,
    \textsuperscript{\rm 2}Amazon Alexa\\
    rsequeir@usc.edu, 
    rosenbrg@amazon.com,
    maminmab@amazon.com,\\
    poliann@amazon.com,
    jonrubin@amazon.com
    
}
\begin{document}
\maketitle

\begin{abstract}
Alexa+ devices with screens display animated icons during conversations to visually reinforce what is being said. The original set of these icons was hand-drawn by designers one at a time, a process that produces polished results but is time-consuming and cannot scale to the range of topics encountered in open-ended conversation. We describe an automated pipeline, deployed in production, that generates these vectorised animated icons from a text description alone and surfaces relevant animations to customers based on their ongoing conversation. The pipeline decomposes icon creation into a sequence of stages (image generation, semantic decomposition, vectorisation, styling, and animation). Its quality is determined at the very first of these, image generation, on which every downstream stage depends. We therefore focus on improving this first and most critical stage. Because the icons are rendered on small display canvases under strict design guidelines, generations are often unsuitable for production. Supervised fine-tuning is not viable, as no dataset of line-art icons exists under these constraints. Instead, we design objective verifiers that automatically flag generations violating the design guidelines, and use them to fine-tune a flow matching image generation model via reinforcement learning from verifiable rewards (RLVR). The fine-tuned model raises the pass rate from 15.7\% to 58.7\% (3.7×) under a VLM judge that is independent of the reward model. Both the trained model and the verifier are deployed in the production pipeline, where the verifier also serves as the automated quality gate before final design review.
\end{abstract}

\section{Introduction}
\label{sec:intro}

\begin{figure}[t]
\centering
\includegraphics[width=\columnwidth]{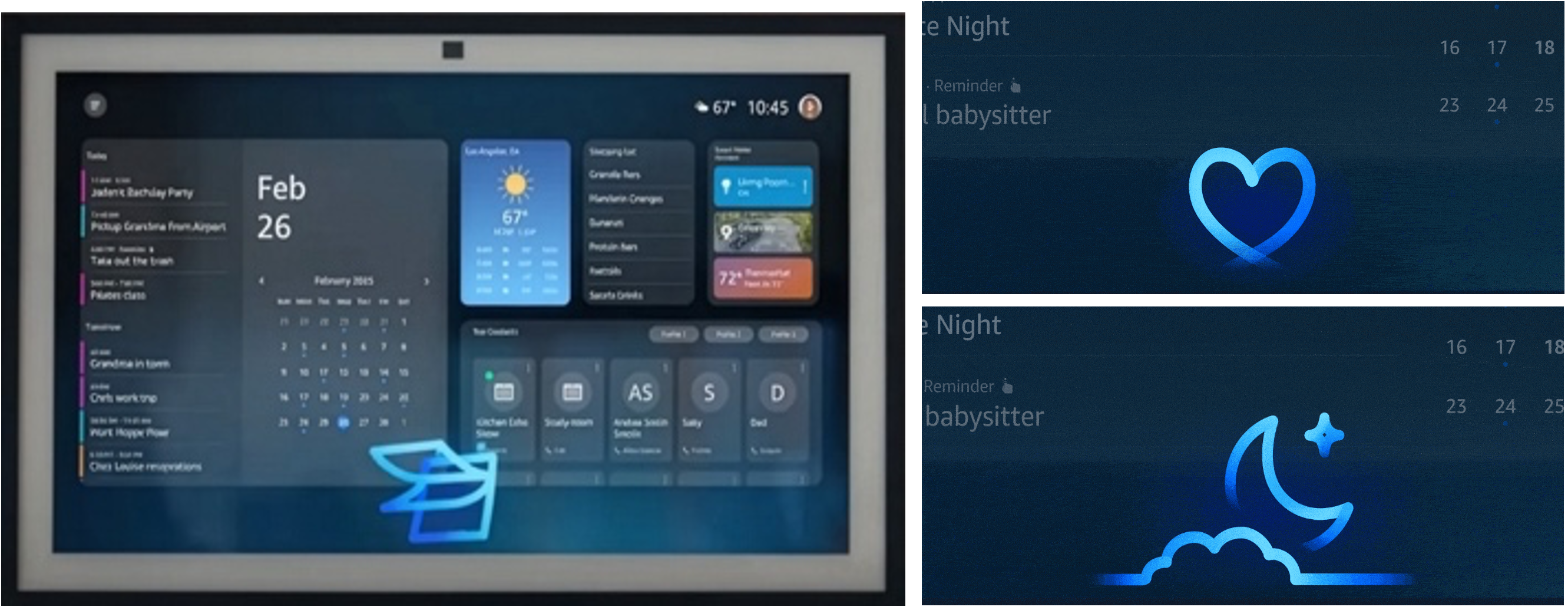}
\caption{Alexicons on an Alexa+ display device. Left: an Alexicon animating on the
home screen. Right: two Alexicons shown at closer scale, rendered in the signature
blue gradient with fluid line-art strokes.}
\label{fig:alexicon-device}
\end{figure}

Conversational assistants on screen-equipped devices communicate through voice, but many
interactions benefit from a visual cue that reinforces what is being said. Alexicons are
the animated line-art icons that serve this role on Alexa+ devices
(Figure~\ref{fig:alexicon-device}): a question about cooking surfaces a skillet, a
weather query brings up an umbrella. Each icon is rendered
in a signature blue gradient with fluid motion, and the original set was hand-drawn by
designers one icon at a time. That process produced polished results but could not keep
pace with the range of topics Alexa+ encounters in open-ended conversation. We built an
automated pipeline that decomposes icon creation into tractable stages: image generation,
semantic decomposition, centreline vectorisation, gradient styling, and procedural 
Lottie \footnote{\url{https://lottie.github.io/}}
animation. The quality of the final icon is determined at the very first stage, and every
downstream stage inherits whatever that first drawing produces. Image generation is the
binding bottleneck.
This paper describes the automated pipeline we use to serve Alexicons in
production traffic, with a focus on the image generation stage and the improvements we made to it.

\begin{figure*}[ht]
\centering
\includegraphics[width=0.8\textwidth]{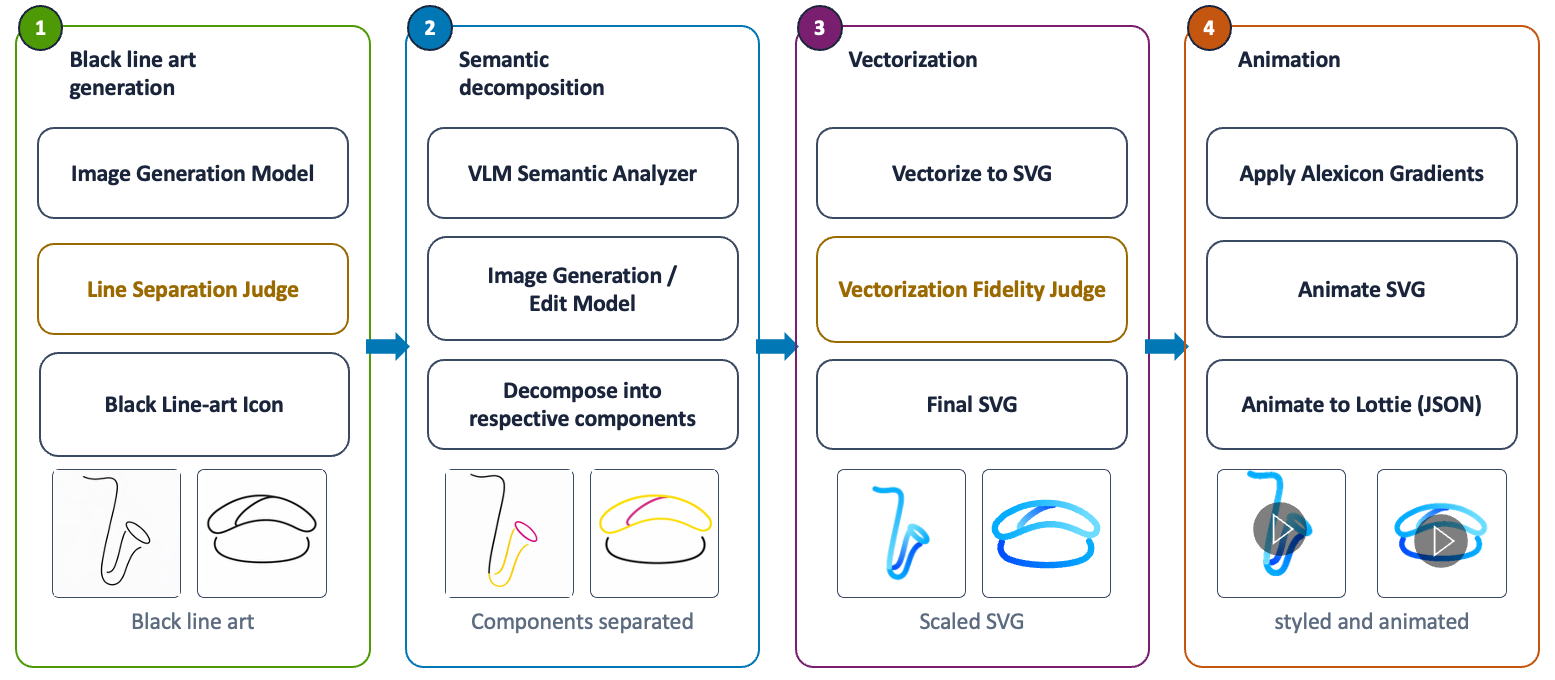}
\caption{The Alexicon production pipeline. Stage~1 generates black line-art from a text
prompt and judges stroke separation quality. Stage~2 uses a VLM to identify semantic
components and an image generation model to decompose the drawing into colour-coded
parts. Stage~3 vectorises each component into centreline SVG paths and verifies
vectorisation fidelity. Stage~4 applies the Alexicon gradient palette, animates the
SVG, and exports a Lottie file for on-device playback.}
\label{fig:pipeline}
\end{figure*}

The initial pipeline relied on a proprietary image generation API. However, it is desirable to have a tailorable model rather than remaining dependent on
frontier image generation services.
FLUX.2 [klein]~\cite{blackforestlabs2025flux2}, a $4$B parameter open-source
flow-matching~\cite{lipman2023flowmatchinggenerativemodeling} model, was identified as a practical candidate given its
permissive licence and undistilled weights. 

Alexicons are rendered on small display
canvases with strict design guidelines that require relatively large stroke widths. When
adjacent contours sit too close in the image generation model's high-resolution output, they merge into
solid shapes once re-rendered at the production stroke width, we call this defect
\emph{stroke crowding}. With proprietary models, we managed this through extensive prompt tuning,
and automated Vision-Language Model (VLM) judges that dropped failing cases at the cost of generating more variations per-icon. 
Other open-source models we evaluated produced the defect far more often and were less reliable.
Out of the box, FLUX.2 [klein] exhibits stroke crowding on roughly $85\%$ of generations.
Generating at lower resolution or controlling stroke width through prompting would
theoretically circumvent the issue, but current image generation models cannot do either reliably while maintaining
the required line-art quality, and vectorization of low-resolution images introduces
artifacts (visual imperfections, jagged shapes, broken contours, etc.) downstream.
Fine-tuning the model to avoid this defect is the natural next step, but supervised
fine-tuning requires a corpus of correct examples, and no public dataset of line-art
icons exists under these rendering constraints.

Reinforcement learning offers an alternative: it can optimise against a
non-differentiable reward without paired data. In this work, we use
Flow-GRPO (Flow based- Group Relative Policy Optimization)~\cite{liu2025flowgrpo} to fine-tune FLUX.2 [klein] against a reward signal that
evaluates the model's own rendered output directly.
We introduce a morphological measurement tool that
flags every region where strokes sit too close. A gradient-boosted classifier trained on
$3{,}836$ hand-labelled violations from $682$ icons separates genuine defects from
spacing the design requires. The trained model and objective verifier are
deployed inside the production pipeline, introducing more objective automated quality
gates into the Alexicon creation process.

\paragraph{Contributions.} Our main contributions are:
\begin{itemize}
\item We present the full automated pipeline for creating vectorised
      animations that are displayed on Alexa+ devices, from text description to on-device
      playback.
\item We present a reward model for stroke crowding that combines a
      stroke-clearance measurement tool with an objective verifier to distinguish genuine
      defects from intended design, requiring no target images and no manual annotation at
      training time.
\item We demonstrate that Flow-GRPO fine-tuning against our proposed reward produces a model that
      generalises the learned spacing behaviour to held-out concepts, evaluated under a VLM judge 
      that is independent of the objective verifier used in the reward model.
\item We deploy the trained model and objective verifier inside the Alexicon generation
      pipeline, where the verifier
      % that shaped the reward also
      serves as the automated quality judge before design review.
\end{itemize}
\section{System Overview}
\label{sec:system-overview}

The Alexicon pipeline (Figure~\ref{fig:pipeline}) turns a short text description into a
finished animation through four stages, each handling a different aspect of the
transformation from concept to on-device usage.
The first stage is image generation. A text-to-image model produces a
high-resolution black-on-white line drawing conditioned on hand-crafted
Alexicon references that anchor the output style: continuous
single-stroke paths, smooth curves, rounded corners, no filled shapes,
and a uniform thin stroke. The generated drawing is then re-rendered at the production
display dimensions with the required stroke width and padding. A VLM judge compares the
original output against this re-render to check whether the icon remains legible without
stroke crowding. Only icons that pass move forward in the pipeline.
The second stage is semantic decomposition. A vision-language model
examines the drawing and identifies its semantic components (a handle, a
blade, a base), outputting a structured description of each part. The
image generation model then re-renders the icon with each
component drawn in a distinct colour, preserving the original geometry
while encoding the part boundaries into the pixel data. The result is a
colour-coded image where each semantic region is visually separated.

Subsequent stages separate the colour-coded image into per-component
raster masks, vectorise each into a centreline SVG, and apply the
Alexicon visual identity. Paths receive colours from a fixed palette of three anchor blues,
a stacking order is inferred for overlapping parts, and opacity fades
soften open stroke terminations. The styled SVG is then turned into a
Lottie animation: a connectivity graph built from path endpoints
determines the drawing order, and composable motion effects (trim-path
reveals, vertex morphing, rotation, distance-based staggering) are
assigned to each part. The final animation is inserted into a template
with the background glow and Alexa voice chrome bar, producing a file
ready for on-device playback.

The pipeline runs end to end without manual intervention, but the first
stage carries the heaviest external dependency:
multiple image generation model calls per icon. Every
downstream stage inherits the quality of the drawing that comes out of
image generation. A line-art with tangled or overcrowded strokes
propagates errors through decomposition, vectorization, styling, and
animation, and no later stage can undo a structural flaw baked into the
geometry of the image.
The rest of the paper focuses on the image generation stage, where stroke crowding
is the dominant quality problem.

\section{Motivation}
\label{sec:motivation}

Generative image models \cite{openai2025gpt4o_image,imagenteamgoogle2024imagen3,labs2025flux1kontextflowmatching,esser2024scalingrectifiedflowtransformers} 
train predominantly on photorealistic images,
where colour, texture, and shading absorb small positional errors. A contour shifted by a few pixels in a
photograph changes almost nothing perceptually. Black line art has none of that cover.
Every stroke is a bare geometric mark, and a shift of a few pixels can push two
contours close enough to merge once the icon is rendered at its production stroke width.
The design constraints that Alexicons must satisfy make the margin even
tighter: the production canvas is small and the required stroke width is large relative
to it, so each stroke consumes a substantial fraction of the available space. Only a
handful of parallel strokes fit side by side with adequate clearance between them, and
most objects need more structure than that limit can carry.

\begin{figure}[t]
\centering
\begin{subfigure}[t]{0.48\columnwidth}
\centering
\includegraphics[width=\linewidth]{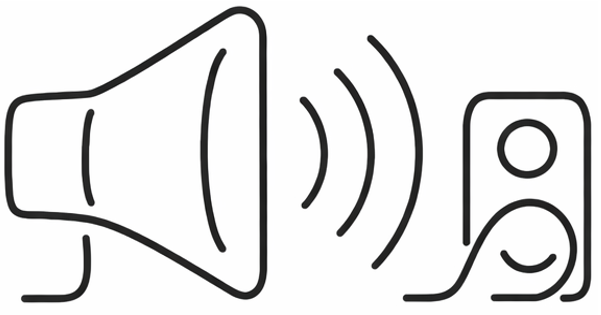}
\caption{Generated at high resolution.}
\label{fig:speaker-hires}
\end{subfigure}
\hfill
\begin{subfigure}[t]{0.48\columnwidth}
\centering
\includegraphics[width=\linewidth]{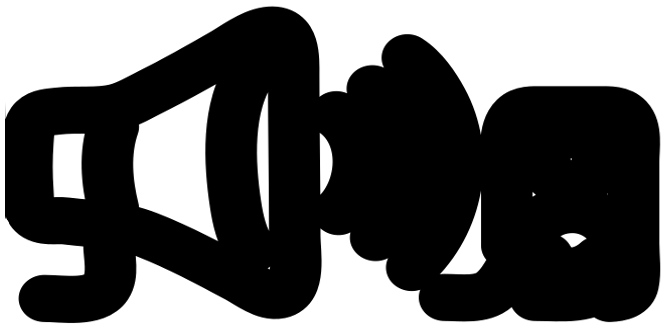}
\caption{Re-rendered at the required stroke width and display size.}
\label{fig:speaker-production}
\end{subfigure}
\caption{Stroke crowding on a generated speaker icon.}
\label{fig:stroke-crowding}
\end{figure}

At the high resolution produced by image generation models, lines appear thin and well
separated (Figure~\ref{fig:stroke-crowding}, left). Stroke crowding only becomes visible
once the drawing is downscaled to the target display size and re-rendered at the required
stroke width (Figure~\ref{fig:stroke-crowding}, right), because contours that had room at
high resolution are now pushed together. The model has no representation of this artifact
in its own output space. Explicit spacing instructions in the prompt do not help. The
model reads the instruction but lacks the sub-stroke-width spatial precision to act on
it.

Automating the detection of stroke crowding is itself harder than it first appears.
A strict geometric measurement can flag every region where two strokes sit closer than
one stroke width, but not every narrow gap is a defect. Bubbles drawn as circles, sharp corners, and joints where two parts of an object meet all produce
sub-stroke-width clearances that a geometric measurement would flag, even though those
tight gaps are necessary to preserve the object's meaning. A measurement tool
that flags all of these indiscriminately would penalise correct drawings alongside
invalid ones, and any reward built on top of it would teach the model the wrong lesson.
Separating genuine crowding defects from spacing that the design requires is a semantic
judgment that geometry alone cannot make.

Since no suitable dataset exists with line-art icons under these rendering
constraints, the model cannot be taught correct spacing through supervised examples.
Instead, it must learn from a signal that evaluates its own output against the rendering
constraint directly.
The next section describes our approach to tackling these problems and the methodology
behind fine-tuning an open-source model to produce well-separated line art under the
Alexicon design constraints.

\section{Method}
\label{sec:method}

Our method fine-tunes an open-source text-to-image model to produce line-art icons
that remain legible at the production stroke width, using a reward signal built
entirely from the rendered output. We begin by constructing a stroke-clearance
measurement tool that flags every region where strokes sit too close on the rendered
drawing. We then describe the verifier dataset assembled to train a classifier on top
of this tool, and the objective verifier that uses that classifier to separate genuine
crowding defects from spacing that the design requires. Finally, we present the full
reward model that combines the verifier's defect count with guards against degenerate
outputs and a concept-fidelity check, forming the training objective for reinforcement
learning.

\begin{figure}[ht]
\centering
\begin{subfigure}[t]{0.31\columnwidth}
\centering
\includegraphics[width=\linewidth]{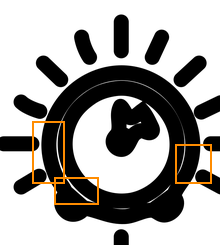}
\caption{Timer dial.}
\label{fig:crowding-narrow}
\end{subfigure}
\hfill
\begin{subfigure}[t]{0.31\columnwidth}
\centering
\includegraphics[width=\linewidth]{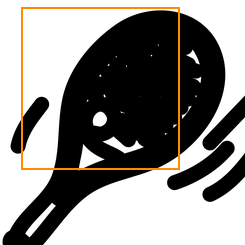}
\caption{Tennis racket.}
\label{fig:crowding-fused}
\end{subfigure}
\hfill
\begin{subfigure}[t]{0.31\columnwidth}
\centering
\includegraphics[width=\linewidth]{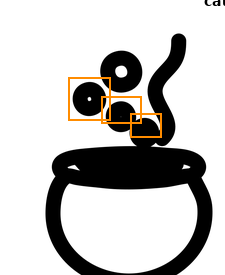}
\caption{Cauldron.}
\label{fig:crowding-holes}
\end{subfigure}
\caption{The three crowding types detected by the measurement tool, with flagged regions
in orange. (a)~Narrow clearance on a timer dial: concentric rings sit too close, leaving
white gaps thinner than one stroke width. (b)~Fused strokes on a tennis racket: the
string grid has merged into a solid black mass. (c)~Enclosed holes on a cauldron: three
of the four bubbles fill into solid dots at the required stroke width.}
\label{fig:crowding-types}
\end{figure}

\subsection{Stroke-clearance measurement tool}
\label{sec:measurement}

The measurement tool reads the binarised drawing, where black pixels are strokes and
white pixels are background, and flags every region where strokes sit too close, fuse
together, or enclose a hole that would fill at the production stroke width. A single
parameter controls the entire tool: the stroke width $w$, and every threshold is
expressed as a multiple of $w$ rather than a pixel constant. 
That anchoring lets the same code run at any render resolution without
per-resolution tuning.
The tool detects three kinds of crowding, discussed in detail below and
each illustrated in Figure~\ref{fig:crowding-types}:
\begin{itemize}
\item \textbf{Narrow clearance} (Figure~\ref{fig:crowding-narrow}): a white gap between
      two adjacent strokes thinner than one stroke width. A morphological closing~\cite{serra1982image} dilates the strokes by
      $w/2$ and then erodes them back, which fills any white gap narrower than $w$
      while leaving wider gaps intact. The filled pixels mark exactly where the
      clearance is insufficient.
\item \textbf{Fused strokes} (Figure~\ref{fig:crowding-fused}): two or more strokes so
      close that the white channel between them has closed entirely and they have merged
      into a single solid region. An exact Euclidean distance
      transform~\cite{felzenszwalb2012distance} of the stroke pixels peaks at $w/2$
      along a clean single stroke, and any region wider than that has merged beyond the
      specified stroke width.
\item \textbf{Enclosed holes} (Figure~\ref{fig:crowding-holes}): small white background
      regions fully surrounded by strokes that fill into solid dots once rendered.
\end{itemize}

\subsection{Verifier dataset}
\label{sec:corpus}

\begin{figure}[t]
\centering
\includegraphics[width=\columnwidth]{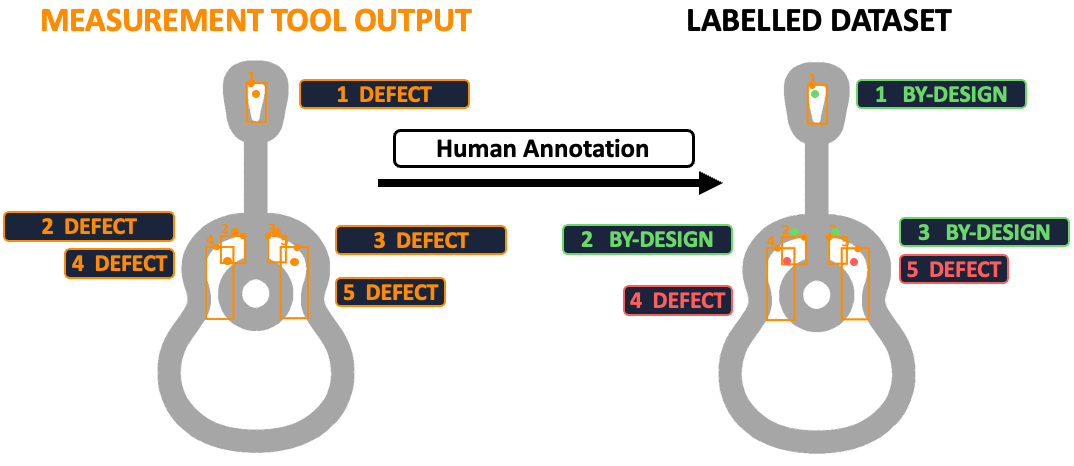}
\caption{From measurement to labelled dataset. Left: the measurement tool
flags five crowding regions on a generated guitar icon, labelling all as potential
defects. Right: after human annotation, regions where tight spacing is structurally
necessary (e.g.\ the neck meeting the head, etc.) are
relabelled \bydesign{}, while regions where contours have genuinely merged remain
\defect{}. The annotated violations form the training data for the objective verifier.}
\label{fig:tool-to-dataset}
\end{figure}

The classifier that sits on top of the measurement tool needs supervised labels at the
violation level, not the icon level, because the reward consumes a count of genuine
defects per icon and the label has to attach to the thing being counted. We assembled
a corpus of $682$ generated icons and ran each through the measurement tool, producing
$3{,}836$ flagged violations at roughly $5.6$ per icon.
Figure~\ref{fig:tool-to-dataset} illustrates how
these violations become training data: the measurement tool flags every crowding region
on a generated icon, and a human annotator then decides for each flag whether the tight
spacing is a genuine \defect{} or structurally necessary and therefore \bydesign{}.
The final corpus contains $1{,}928$ \defect{} against
$1{,}908$ \bydesign{}. The corpus is split by icon, not by violation, into $409$ train,
$137$ validation, and $136$ test icons so that siblings of the same violation never
appear on both sides of the split boundary.

\subsection{Objective verifier}
\label{sec:verifier}

The measurement tool flags every region where strokes sit too close, but it does so
indiscriminately. A scissor's crossed blades, a wristwatch's band meeting its case, and
the rim of a bubble all produce the same geometric signature as an accidental tangle. The
measurement supplies recall; a classifier on top of it supplies precision.

A histogram gradient-boosting ensemble over nine features performs the classification,
deciding for each flagged region whether it represents a genuine defect (\real{}) or
intended design (\intended{}). The classifier uses nine geometric features extracted from
the measurement tool's output, grouped into per-violation properties (how far the crowding
persists, which type it is, the region's size and shape), icon-level counts (how many
crowding regions the drawing has overall), and within-icon context (how this region ranks
among the icon's flags and how far it sits from the nearest other flag). The full feature
list and the rationale behind the selection are in Appendix
Table~\ref{tab:features}. Three properties of the data guided the choice of classifier.
The signal is interactive: a long extent means one thing in a crowded icon and another in
a clean one, and trees capture these interactions without hand-built cross terms. The data
is small and tabular at $3{,}836$ rows over nine columns, the regime where boosted trees
match or beat tuned neural models~\cite{grinsztajn2022tree}.

On the $769$ violations of the $136$ held-out test icons, the classifier reaches a macro
$F_1$ of $0.685$, with \real{} precision $0.697$ and recall $0.646$. We also built two
vision-language alternatives under the same protocol: one replacing the geometric features
with VLM-elicited perception features and refitting the same classifier, the other asking
a VLM for the verdict directly. Both landed in the same accuracy range, but the
geometric classifier is the clear choice for our use case. It is a deterministic
classifier that runs on CPU, requiring no GPU endpoint and no network round trip per
image. It returns the same verdict for the same input every time, where a prompted model
drifts with phrasing and sampling temperature. These are essential for verifiers that run on every sample of every training batch as part of a reward model.

\subsection{Reward model}
\label{sec:reward}

The objective verifier labels each flagged region as a genuine defect or intended
design. The reward model counts those genuine defects per icon, but that count alone
is not a safe reward. A model optimised purely to minimise defect count can draw two clean strokes
of the wrong object and score perfectly: no crowding, no defects, no resemblance to the
requested concept. The reward has to credit the right object drawn with well-separated
strokes, penalise outputs that are not line art at all, and keep a usable gradient on
failed samples so that the model always has a direction to improve in.
We assemble the reward from three components, each addressing a distinct failure mode,
and combine them through a concept-gated sum. For a generated image $x$ and target
concept $c$:

\paragraph{Degradation score $P$.}
Early training runs revealed that the model learned to minimise crowding defects by
drawing less: producing near-empty canvases or shrinking the icon into a small corner
where the measurement tool finds nothing to flag. The degradation score counters
this by computing the ratio of black pixels to the total image area and penalising
any output that falls below a stroke-density floor. Small or sparse drawings are
particularly dangerous because the stroke-clearance measurement tool operates on the
rendered strokes, and a drawing that occupies only a fraction of the canvas will
register few or no crowding violations regardless of its actual quality. A second
term catches chromatic leakage: colour appearing in what should be a black-and-white
drawing. Both terms are graded by severity:
\begin{gather}
P(x) = p_{\text{blank}}(x) + p_{\text{color}}(x), \\
p_{\text{blank}}(x) = -\lambda_b \max\!\bigl(0,\; 1 - \tfrac{\rho}{\rho_0}\bigr), \\
p_{\text{color}}(x) = -\lambda_c \min\!\bigl(1, \tfrac{\gamma}{\gamma_0}\bigr),
\end{gather}
where $\rho$ is the black-pixel ratio of the image, $\rho_0$ is the stroke-density
floor below which the score activates, and $\gamma$ is the coloured-pixel fraction.
The density floor $\rho_0$ was set to the average black-pixel ratio computed across
the generated images already collected for the verifier dataset, giving a natural baseline for how much stroke content a
well-formed icon should contain. The blank term ramps linearly from $0$ when the
drawing meets that floor to $-\lambda_b$ at an empty canvas. Grading rather than
thresholding came from a measured
failure: an earlier hard threshold left no gradient inside already-degenerate
GRPO groups (the set of samples generated for the same concept in one training
step),

\paragraph{Concept-fidelity score $F$.}
The degradation score stops the model from producing near-empty or coloured outputs,
but it does not check whether the drawing actually depicts the requested concept. A
model that draws a well-sized, black-and-white icon of the wrong object would pass the
degradation check and score well on stroke crowding. The concept-fidelity score closes that
gap by using a vision-language model as a semantic judge. Pixel-level statistics cannot
determine whether a drawing of a skillet actually looks like a skillet; that judgment
requires a model that understands visual semantics. A single VLM call per sample
produces three binary flags: concept match $m$, which checks whether the drawing
depicts the requested object; a minimalism check $q_{\min}$, which verifies that
the drawing uses clean, sparse strokes rather than dense or filled regions; and a
crowding check $q_{\mathrm{crd}}$, which flags drawings where strokes are visibly
tangled. Concept match gates the other two flags: a wrong object zeroes out the
entire raw score regardless of how tidy the rendering is:
\begin{gather}
f(x) = w_m\,m + m\bigl(w_q\,q_{\min} + w_u\,q_{\mathrm{crd}}\bigr), \\
F(x) = \kappa\,\bigl(f(x) - f_0\bigr),
\end{gather}
with $w_m = 0.6$, $w_q = w_u = 0.2$, $f_0 = 0.4$, and $\kappa = 5.0$.

\paragraph{Stroke-crowding score $\sigma$.}
The stroke-crowding score is the primary reward signal: it translates the objective
verifier's defect count into a training gradient that pushes the model toward
well-separated strokes. The degradation and concept-fidelity scores exist to prevent
the model from gaming this signal by drawing less or drawing the wrong object. The score converts the defect count directly rather than using a rate. An earlier
rate-based version collapsed during training, producing zero group variance in
$41.4\%$ of iterations. Analysis showed that samples with different absolute defect
counts but the same defect-to-flag ratio scored identically under the rate
formulation, flattening the gradient. Switching to raw count preserves the difference
between two defects and five, eliminating the problem:
\begin{gather}
s(x) = \tfrac{1}{1+n}, \\
\sigma(x) = \underbrace{w_\sigma(2s-1)}_{\text{crowding}}
          - \underbrace{\min\!\bigl(\sigma_{\max},\, w_e\,\pos{n-\tau}\bigr)}_{\text{excess ramp}},
\end{gather}
with $w_\sigma = 1.0$, $\tau = 3$, $w_e = 0.5$, and $\sigma_{\max} = 2.0$. The first
term runs from $+1$ at zero defects toward $-1$ as $n$ grows, but it flattens past three
defects. The excess ramp restores a gradient in the tail, subtracting $0.5$ per defect
beyond $\tau$, which keeps $\sigma$ bounded in $(-3, +1]$ and strictly
decreasing.

\paragraph{Combined reward.}
The final reward sums the three scores but gates two of them on the concept-match
flag $m$ produced by the VLM judge in the concept-fidelity score:
\[
R(x,c) = P(x) + w(m)\,\sigma(x) + \tilde F(x,c),
\]
When the VLM confirms a correct concept ($m=1$), the stroke-crowding score enters at
full strength and the fidelity score contributes positively. When the concept is wrong
($m=0$), the fidelity is floored at $-1.0$ and the stroke-crowding score is attenuated
to $30\%$. The attenuation is deliberate. Setting the stroke-crowding weight to zero on
wrong-concept samples would flatten those GRPO groups entirely, removing the gradient that
tells the model a clean wrong drawing is closer to the goal than a messy one. Keeping
full strength would do the opposite: rewarding well-spaced drawings of the wrong
object. The $30\%$ weight balances both risks, leaving enough signal for the model to
improve stroke quality even on failed concepts while ensuring that a concept fix is
required to escape the negative reward range.

\section{Experimental Setup}
\label{sec:experimental-setup}

In this section we describe the reinforcement learning configuration
used to fine-tune the image generation model against the reward signal, including the
training concept list (prompts), the model configuration, and the training reward trajectory.

\subsection{Reinforcement learning configuration}

\paragraph{Training concept list.}
Reinforcement learning needs concepts that produce disagreement between samples of the
same concept. A concept that every sample already draws well, and one that no sample can
draw at all, both yield zero within-group variance and drop out of the update. The
training concept list is built around this constraint. It contains $592$ concepts across
four tiers of escalating compositional demand: three geometric primitives in mixed spatial
relations ($80$ concepts), one free-form abstract shape ($125$), one real-world object
($225$), and one object shown mid-action ($162$). The last tier is hardest because an
action has to be conveyed through outline alone. The list is split $474$ train and
$118$ held-out, stratified within each tier at $80/20$.

Each entry stores a bare concept description. A fixed structured template wraps the
description at build time, following the FLUX.2 prompting guide where every line-art
requirement is phrased positively: \textit{``each edge outlined once''} carries what \textit{``no double
lines''} would have, and \textit{``wide airy spacing between strokes''} replaces \textit{``do not draw lines
close together.''}

\paragraph{Model and training.}
Flow-GRPO~\cite{liu2025flowgrpo} is the algorithm used to fine-tune FLUX.2 [klein] 4B model
($4$B rectified-flow transformer with a $4$B Qwen3 text encoder, released under
Apache-2.0) against the combined reward. GRPO eliminates the need for both a separate critic network and reference images by
estimating advantages from within-group comparisons, which fits our setting where no
target image exists. Flow-GRPO extends it to continuous-time flow-matching models like
FLUX. Two properties decided the model choice. It is undistilled, so training and evaluation
share one sampler configuration with a real classifier-free guidance scale. The permissive
licence allows fine-tuning and serving inside a production pipeline.
A Low-Rank Adaptation (LoRA) adapter \cite{hu2021loralowrankadaptationlarge} is trained on top of the frozen base weights, keeping the base model's
learned representations intact while adapting its behaviour to the Alexicon constraints.
The adapter targets the
attention projections of both the image and text streams and the fused projection in the
single-stream blocks~\cite{liu2025flowgrpo}. Training is strictly on-policy with one inner epoch per rollout.
The full hyperparameter configuration is in Appendix Table~\ref{tab:hparams}.

\begin{figure*}[ht]
\centering
\includegraphics[width=\textwidth]{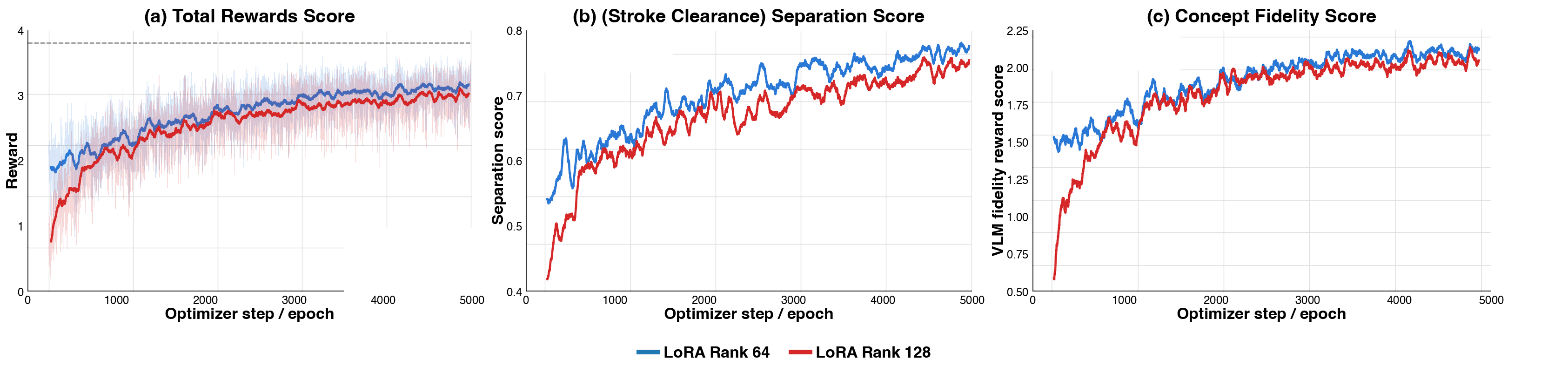}
\caption{Training reward over $5{,}000$ optimiser steps for LoRA rank
$64$ and rank $128$. (a)~Combined reward: both configurations improve steadily, with
rank $64$ reaching a higher plateau; the dashed line marks the theoretical maximum
($R\!=\!4.0$). (b)~Line separation score: the stroke-crowding component of the reward
rises from ${\sim}0.48$ to ${\sim}0.78$, showing that the model learns to space strokes
farther apart over training. (c)~Concept fidelity score: the VLM-based fidelity
component rises from ${\sim}0.5$ to ${\sim}2.0$, confirming that depiction accuracy
improves alongside stroke separation rather than trading off against it.}
\label{fig:training-curves}
\end{figure*}

Figure~\ref{fig:training-curves} shows the training reward trajectory
over $5{,}000$ optimiser steps for two LoRA rank configurations. The combined reward
(Figure~\ref{fig:training-curves}a) on the $474$ training concepts rises steadily from
a baseline of $1.71$ to $2.85$, a $66.7\%$ relative improvement. Decomposing the reward
into its components confirms that the model learns useful signals on both fronts: the
line separation score (Figure~\ref{fig:training-curves}b) climbs from ${\sim}0.48$ to
${\sim}0.78$, showing that the model progressively spaces its strokes farther apart,
while the concept fidelity score (Figure~\ref{fig:training-curves}c) rises from
${\sim}0.5$ to ${\sim}2.0$, indicating that the model maintains and improves depiction
accuracy as it learns to separate strokes. Both components improve in tandem rather than
trading off against each other. The held-out reward on the $118$ reserved training
concepts tracks the same trajectory, peaking at $2.71$ ($58.5\%$ relative improvement)
before beginning to plateau. We select the weights at that peak to
represent the trained model in all subsequent evaluation.
\section{Results}
\label{sec:results}

We evaluate the objective verifier on its held-out test set, then measure
the trained model's image generation quality under an independent VLM judge on a evaluation concept list.

\subsection{Objective Verifier Performance}
The objective verifier reaches a macro $F_1$ of $0.685$ on the $769$ violations of the
$136$ held-out test icons, with \defect{} precision $0.697$ and recall $0.646$.
Table~\ref{tab:verifier} shows the confusion matrix. The $107$ false \defect{} predictions
are the costliest errors: each one penalizes a correctly drawn region and feeds wrong
gradient into the reward. A precision of $0.70$ is not a clean oracle, and the residual
error transfers into the reward as label noise on the defect count. Stroke crowding is
an inherently subjective judgment: the boundary between a tight-but-acceptable crossing
and a genuine defect depends on the object being depicted, and annotators disagree on
ambiguous cases. This subjectivity is reflected in the label noise. Tightening the boundary would
require a larger annotated corpus, stricter annotation guidelines that codify
per-junction-type rules, and multi-annotator agreement to reduce bias.

\begin{table}[t]
\centering
\renewcommand{\arraystretch}{1.3}
\begin{tabular}{@{}l c c@{}}
\toprule
& pred.\ \defect{} & pred.\ \bydesign{} \\
\midrule
true \defect{}     & $246$ & $135$ \\
true \bydesign{} & $107$ & $281$ \\
\bottomrule
\end{tabular}
\caption{Confusion matrix on the $769$ violations of the $136$ held-out test icons,
positive class \defect{}. The $107$ false \defect{} predictions are the cell the downstream
reward weights most heavily.}
\label{tab:verifier}
\end{table}

\subsection{Image Generation Evaluation}
\paragraph{Evaluation concept list.}
The evaluation concept list contains $1{,}000$ concepts, fully disjoint
from the training concept list. The evaluation uses
the same fixed structured prompt template as training so that any difference in pass rate
reflects a change in the model's generation behavior, not a difference in how the concept
was presented to the model.

\paragraph{Evaluation method.}
The trained model was optimized against the proposed reward model, which depends
primarily on the objective verifier and its underlying measurement tool. If the model
learned quirks of those components rather than genuinely improving stroke separation,
evaluating with the same tools would not catch the difference.
Hence, the evaluation uses an independent VLM judge (Claude Opus 4.6) capable of
comparing visual details across images. It receives two images: the
original generated image, and a re-render of the same centerline at the production
stroke width. The judge is asked whether lines that are distinct in the original stay
distinct once rendered at the production stroke width. Output is a simple, PASS or
FAIL.

Each concept in the evaluation list is generated once by the base model and once by the
trained model, with every other setting (template, sampler, judge, and the script that
re-renders centrelines at the production stroke width) held identical. Each generation
is judged once with no retries, so the reported number is the image generation pass
rate: the fraction of icons that clear the stroke-crowding gate on a single generation.
Because the same concept drives both generations, each concept yields a paired verdict,
and the significance test is McNemar's exact two-sided binomial on the discordant
pairs~\cite{mcnemar1947note}.

\subsection{Image Generation Quality}
\begin{figure}[t]
\centering
\includegraphics[width=\columnwidth]{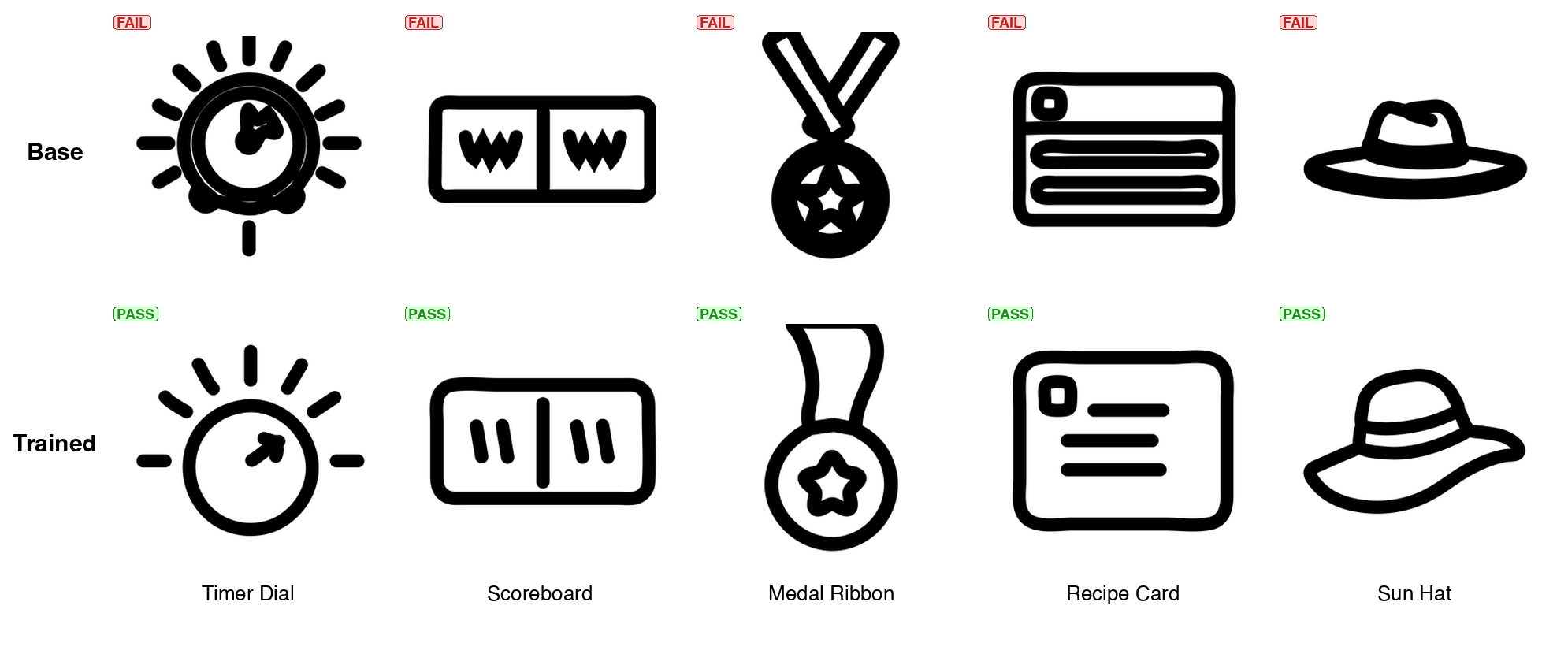}
\caption{Base model (top) vs trained model (bottom) on five evaluation
concepts, rendered at the production stroke width. In all five, the base outputs have strokes
merge into solid masses, while the trained model learned to simplify
internal structure and space contours apart.}
\label{fig:comparison-pass}
\end{figure}

Under this evaluation, the trained model raises the pass rate from
$15.7\%$ to $58.7\%$ (Table~\ref{tab:main}), a $3.7\times$ improvement. Roughly three
in five generations
now clear the stroke-crowding gate, against one in six before.
Figure~\ref{fig:comparison-pass} shows representative examples. The
base model draws fine detail that collapses into solid masses at the production stroke
width: scoreboard digits fuse, medal ribbon edges merge, and recipe card lines
disappear. The trained model has learned to simplify internal structure and widen gaps
between contours, producing icons that remain legible after re-rendering. A larger panel
of base and trained outputs is provided in Appendix~\ref{sec:additional-examples}.

The paired contingency table (Table~\ref{tab:confusion}) breaks the numbers into their
four cells. The base model failed and the trained model passed on $530$ generations
against $40$ in the reverse direction, a thirteen-to-one flip ratio. McNemar's test~\cite{mcnemar1947note} checks whether this asymmetry reflects a
genuine shift in model behaviour or just concept-to-concept variation: if training had no
effect, flips toward passing and flips toward failing should be roughly equal among the
$570$ discordant pairs. The two-sided $p$-value is $3.3\times10^{-94}$, far below the
conventional $\alpha = 0.05$ threshold, confirming that the trained model's higher pass
rate is a systematic effect of fine-tuning rather than noise in the evaluation set.

\begin{table}[t]
\centering
\renewcommand{\arraystretch}{1.3}
\begin{tabular}{@{}l r r r r@{}}
\toprule
& $n$ & Base & Trained & $\Delta$ \\
\midrule
Icon concepts & $1{,}000$ & $15.7\%$ & $58.7\%$ & $+43.0$ pp \\
\bottomrule
\end{tabular}
\caption{Image generation pass rate on the $1{,}000$-concept evaluation list under the
independent VLM judge, one generation per concept. Base is FLUX.2 [Klein] base 4B model; trained
is the trained LoRA weights merged into the base weights. Both arms use same settings and are judged on centerline re-renders to the
production stroke width.}
\label{tab:main}
\end{table}

\begin{table}[ht]
\centering
\renewcommand{\arraystretch}{1.3}
\begin{tabular}{@{}l r r@{}}
\toprule
& Trained PASS & Trained FAIL \\
\midrule
Base PASS & $157$ & $40$ \\
Base FAIL & $530$ & $273$ \\
\bottomrule
\end{tabular}
\caption{Paired verdicts on the $1{,}000$-concept evaluation list. The off-diagonal cells
carry the test: $530$ concepts the base failed and the trained model passed, against
$40$ in the reverse direction.}
\label{tab:confusion}
\end{table}

\section{Analysis}
\label{sec:analysis}

\subsection{Prompt Engineering vs Fine-Tuning}
Each concept in the evaluation list is generated under two prompt
variants: a \emph{plain} prompt that describes the icon without mentioning stroke
placement, and a \emph{separation} prompt that adds explicit instructions to keep
strokes widely spaced and avoid touching lines. Both the base model and the trained
model are evaluated under each variant, with McNemar's exact two-sided
binomial~\cite{mcnemar1947note} testing significance on the discordant pairs.

Adding the separation instruction to the base model moves the pass rate from $15.7\%$
to $13.7\%$ ($p = 0.159$), a change indistinguishable from noise. The model reads the
instruction, but it lacks the spatial precision to act on it. The trained model with the
plain prompt scores $58.7\%$, a $45.0$ percentage-point gap over the prompted base model
($p = 1.3\times10^{-103}$), confirming that the improvement is in the weights rather
than the phrasing. Adding the separation instruction to the trained model produces
$56.6\%$ ($p = 0.291$): the instruction is redundant once the behaviour is learned.

\subsection{Failure Mode Analysis}
To understand how uniformly the improvement distributes, we group the $1{,}000$
evaluation concepts by the compositional complexity of the object they describe: simple
shapes with few strokes (e.g.\ a canoe, a crescent moon), moderately detailed objects
(e.g.\ a coffee mug with handle and rim), and dense compositions with many overlapping
parts (e.g.\ a pipe wrench with jaws, teeth, and graduations). Every complexity group
improves by at least $33$ percentage points, and the ordering is preserved: concepts the
base model found hardest remain the harder ones after training. Dense compositions have
less clearance to spare at the production stroke width, so the absolute pass rate stays
lower even after fine-tuning.

\begin{figure}[t]
\centering
\includegraphics[width=\columnwidth]{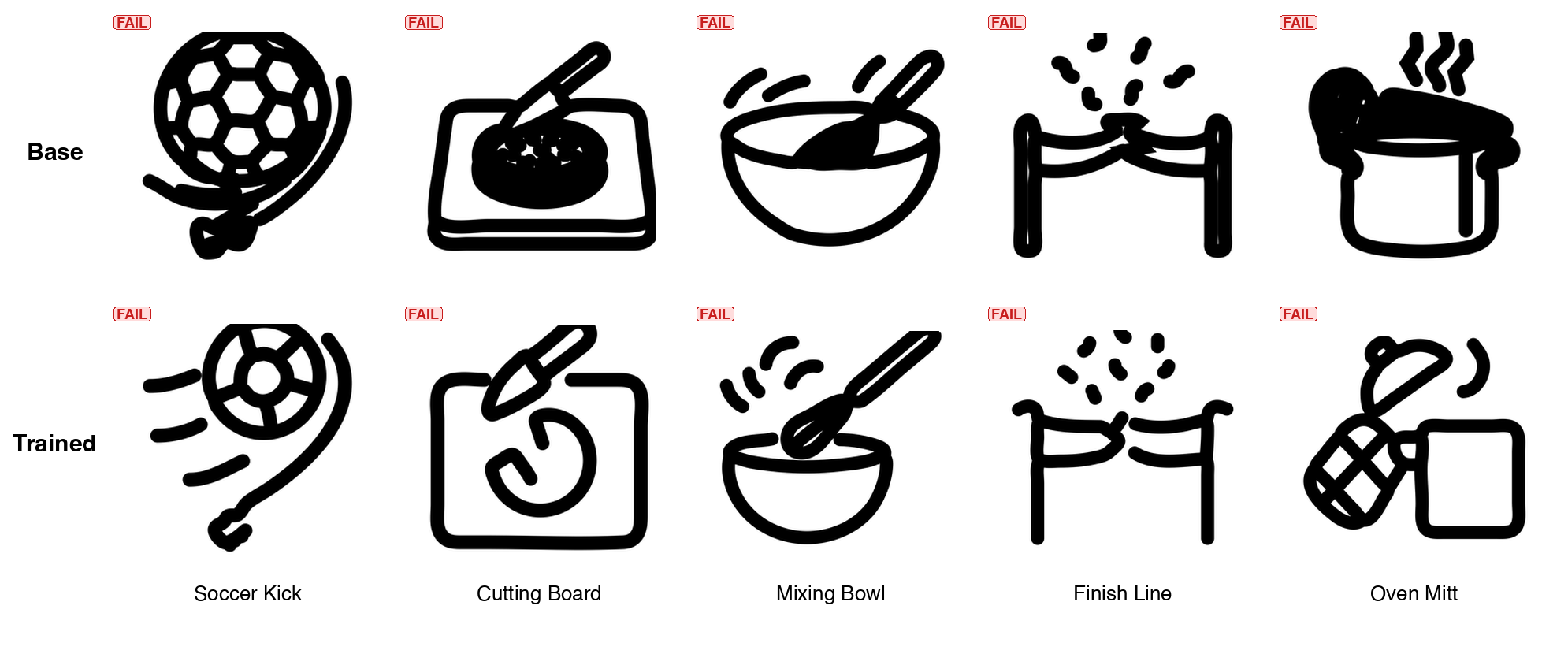}
\caption{Some concepts where both the base model (top) and
trained model (bottom) fail. Failures fall into two categories: local junction merges
where contours sit too close (cutting board, mixing bowl), and concepts whose visual
complexity exceeds what the canvas can separate at the production stroke width (soccer
kick, finish line, oven mitt).}
\label{fig:comparison-fail}
\end{figure}

To understand how training changes the nature of failures, we analyse the one-sentence
rationales the VLM evaluation judge writes for each verdict. Of the base model's
failures, $60.5\%$ were whole regions collapsing to a solid mass at the production
stroke width. That mode falls to $31.9\%$ of the trained model's residual failures.
Training removed the gross failure: a subassembly lost entirely to a filled shape. What
remains is local: neighbouring contours merging at a single junction ($53.8\%$ of
residual failures), a lid separator touching a container outline or a handle meeting a
body edge. The trained model fails on drawings that are nearly right at one point, where
the base model failed on drawings that were illegible.
Figure~\ref{fig:comparison-fail} shows five concepts where both models
fail. The cutting board and mixing bowl are junction failures: the board's edges merge
where they meet the food, and the whisk strokes crowd against the bowl rim. The soccer
kick illustrates a different problem: the trained model simplifies the ball until it no
longer reads as a soccer ball, and the line meant to represent a foot is unrecognisable.
The oven mitt and finish line are complexity failures: both objects require too many
strokes to represent legibly, and neither model produces a drawing that conveys the
concept.

A ceiling remains that no reweighting of stroke placement can escape. Some concepts
need more strokes than the canvas has clearance for at the production stroke width.
Getting past that bound requires letting the model simplify the concept, a different
objective from the one we trained.
\section{Conclusion}
\label{sec:conclusion}

We presented the Alexicon pipeline and showed that an objective verifier
built from morphological stroke-clearance measurement and a gradient-boosted classifier
can serve as an RL reward signal to fine-tune FLUX.2 [klein] for line-art icon generation,
raising the pass rate from $15.7\%$ to $58.7\%$ under an independent VLM judge. Both the
fine-tuned model and the verifier are deployed in production.

Three directions remain open. First, the verifier's macro $F_1$ of $0.685$ introduces
label noise into the reward; a larger, multi-annotator corpus with stricter per-junction
annotation guidelines would sharpen the defect boundary and propagate cleaner gradients
into training. Second, a structural ceiling persists for concepts whose visual
complexity exceeds what the canvas can separate at the production stroke width. Crossing
that ceiling requires a reward that credits concept simplification, letting the model
omit or merge strokes when faithful reproduction would guarantee crowding, along with a
curated dataset of complex, high-stroke-count concepts to train and evaluate against. Third, the
current model is trained against a single set of design constraints. Training against
variable render scales and stroke widths simultaneously would test whether the model can
learn multiple design principles at once, increasing the pipeline's flexibility as
design requirements evolve.

\bibliography{references}

\clearpage
\renewcommand{\thesection}{\Alph{section}}
\setcounter{section}{0}
\renewcommand{\thetable}{\Alph{section}.\arabic{table}}

\section*{Appendix}

\refstepcounter{section}
\subsection*{A.~Training Configuration}
\label{sec:appendix-hparams}
\setcounter{table}{0}

Table~\ref{tab:hparams} lists the full hyperparameter configuration used for training.

\begin{table*}[htbp]
\centering
\renewcommand{\arraystretch}{1.3}
\begin{tabular}{@{}l l@{}}
\toprule
Setting & Value \\
\midrule
Base policy & FLUX.2 [klein] 4B model ($4$B DiT $+$ $4$B Qwen3 encoder) \\
Trained parameters & LoRA only; rank $64$, alpha $128$, scaling $2.0$ \\
Optimiser & AdamW, $\beta = (0.9, 0.999)$, $\epsilon = 10^{-8}$, weight decay $10^{-4}$ \\
Learning rate & peak $5\times10^{-5}$, $50$-step linear warmup, cosine to $10\%$ of peak \\
Gradient clipping & global norm $0.5$ \\
Precision & \code{bfloat16}, TF32 matmuls, gradient checkpointing on \\
Parallelism & DeepSpeed ZeRO-2, $8\times$ H200, single node \\
Resolution & $512\times512$, guidance scale $4.0$ \\
Group size $k$ & $8$ samples per concept \\
Batch per optimiser step & $128$ images ($16$ unique concepts $\times$ $8$ samples) \\
Rollout denoising steps & $10$ \\
KL coefficient $\beta$ & $0.08$ against the adapter-disabled reference \\
PPO clip range & $10^{-3}$; advantage clipped to $\pm 5$ \\
Iterations & $5{,}000$ optimiser steps, checkpoint every $50$ \\
\bottomrule
\end{tabular}
\caption{Shipped training configuration. Training occupied eight H200 GPUs on one node.}
\label{tab:hparams}
\end{table*}

\refstepcounter{section}
\subsection*{B.~Verifier Features}
\label{sec:appendix-features}
\setcounter{table}{0}

\begin{table*}[h!]
\centering
\renewcommand{\arraystretch}{1.3}
\begin{tabular}{@{}l l l@{}}
\toprule
Group & Feature & What it captures \\
\midrule
\multirow{5}{*}{Per-violation}
 & Contour persistence      & how far the crowding runs along the contour, in stroke widths \\
 & Crowding type            & which of the three types (narrow, fused, enclosed) \\
 & Region area              & absolute size of the flagged region \\
 & Region aspect ratio      & elongated channel vs.\ compact junction \\
 & Region height            & height of the flagged region in stroke widths \\
\midrule
\multirow{2}{*}{Icon-level}
 & Fusion count             & how many fused-stroke regions the drawing has overall \\
 & Crowding count           & how many narrow-clearance regions the drawing has overall \\
\midrule
\multirow{2}{*}{Within-icon}
 & Severity rank            & rank of this region among the icon's flags by persistence \\
 & Nearest-neighbour distance & distance to the nearest other flagged region, in stroke widths \\
\bottomrule
\end{tabular}
\caption{The nine features used by the objective verifier, grouped by scope. Five describe
the flagged region itself, two summarise the icon's overall crowding, and two relate a
region to its siblings. No feature reads an absolute canvas position, leaving the
classifier invariant to translation and reflection.}
\label{tab:features}
\end{table*}

The objective verifier classifies each flagged region using nine geometric features
extracted from the measurement tool's output. Table~\ref{tab:features} lists them
grouped by scope.

Persistence carries more signal than severity: an intended crossing touches and
separates within about one stroke width, while an accidental tangle smears along the
contour for many. That is why contour persistence is in the model and the worst-point
clearance is not. The two icon-level counts supply context a human annotator uses
without thinking: one crossing in an otherwise clean five-stroke icon is almost certainly
a design feature, while the same crossing in an icon with fourteen other flagged regions
is part of a general mess. No feature reads an absolute canvas position, so the feature
vector is invariant to translation and to horizontal or vertical reflection.

\clearpage

\refstepcounter{section}
\subsection*{C.~Additional Generation Examples}
\label{sec:additional-examples}

We present a broader sample of base and trained model outputs across the
evaluation concept list, grouped by outcome. All icons are rendered at the production
stroke width, and verdicts are from the independent VLM judge.

\begin{figure*}[h!]
\centering
\includegraphics[width=0.70\textwidth]{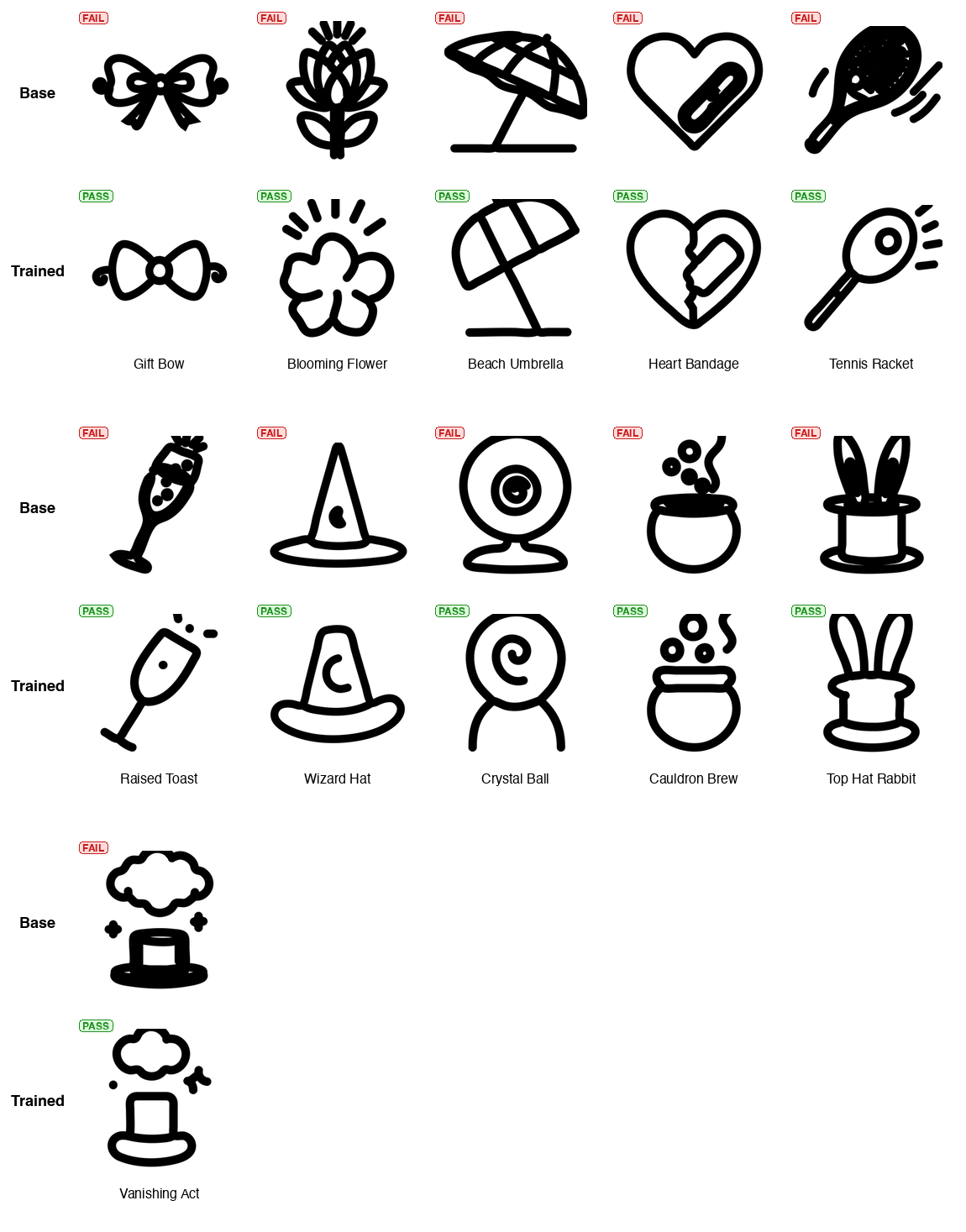}
\caption{Concepts where the base model (top of each pair) fails and the
trained model (bottom) passes. The trained model consistently simplifies internal
structure and spaces contours apart, from simple shapes (gift bow, blooming flower) to
complex compositions (cauldron brew, top hat rabbit, vanishing act).}
\label{fig:appendix-trained-wins}
\end{figure*}

\paragraph{Trained model wins.}
Figure~\ref{fig:appendix-trained-wins} shows $11$ concepts where the base model fails
and the trained model passes. The pattern is consistent across complexity: the base model
draws fine internal detail that collapses into solid masses, while the trained model
simplifies structure and widens gaps. Simple concepts like the gift bow and blooming
flower gain clean contours, and complex ones like the crystal ball and top hat rabbit
become legible where the base produced filled shapes.

\begin{figure*}[h!]
\centering
\includegraphics[width=0.70\textwidth]{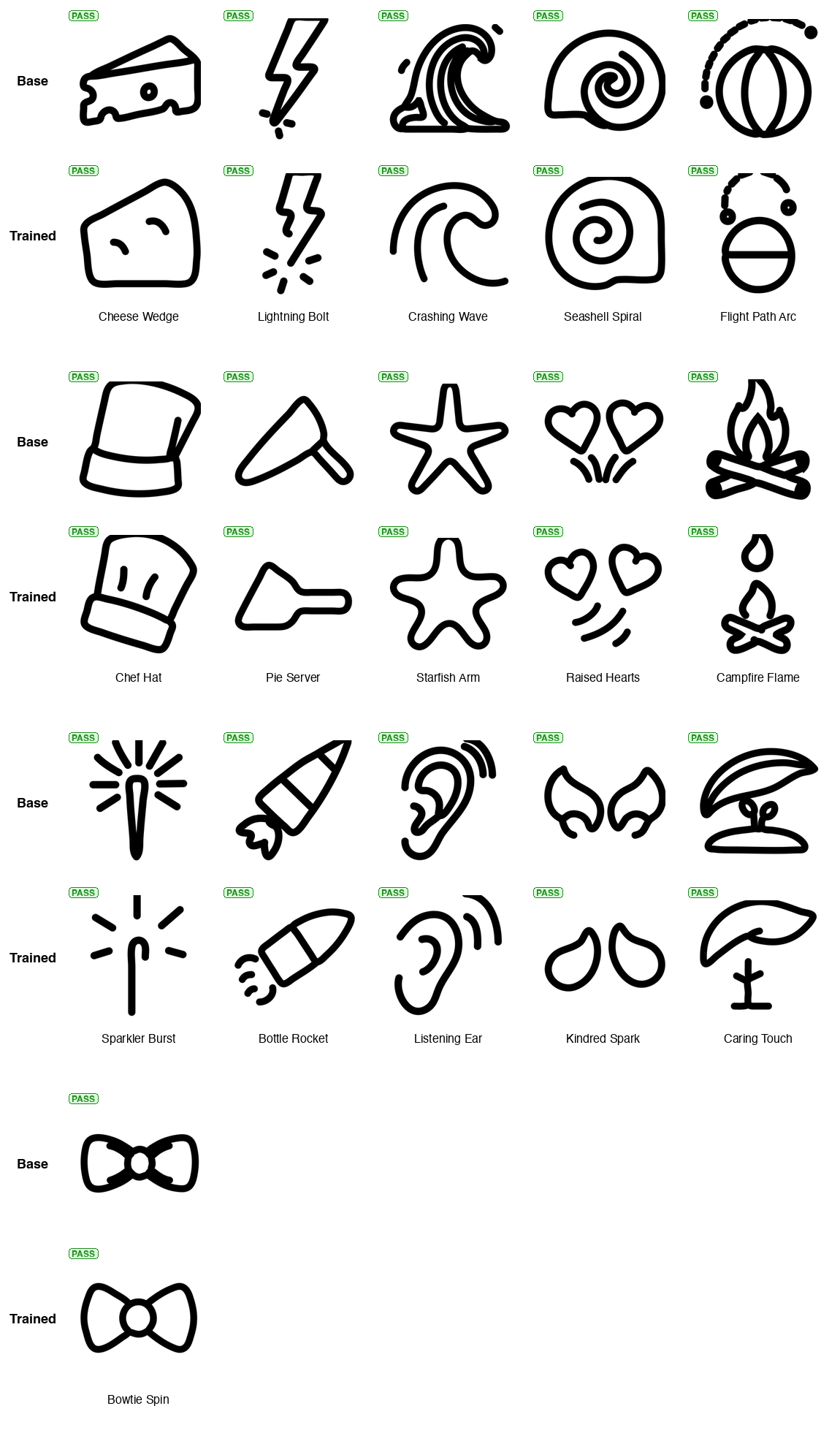}
\caption{Concepts where both the base model (top of each pair) and the
trained model (bottom) pass. These are typically simpler shapes or concepts with
naturally wide spacing, confirming that fine-tuning does not degrade quality on concepts
the base model already drew well.}
\label{fig:appendix-both-pass}
\end{figure*}

\paragraph{Both models pass.}
Figure~\ref{fig:appendix-both-pass} shows $16$ concepts where both models pass. These
are predominantly simpler shapes with fewer strokes (cheese wedge, lightning bolt,
starfish arm) or concepts whose structure naturally leaves wide gaps between contours
(crashing wave, seashell spiral). The trained model produces visually similar outputs on
these concepts, confirming that fine-tuning preserves quality on icons the base model
already handled well.

\begin{figure*}[h!]
\centering
\includegraphics[width=0.70\textwidth]{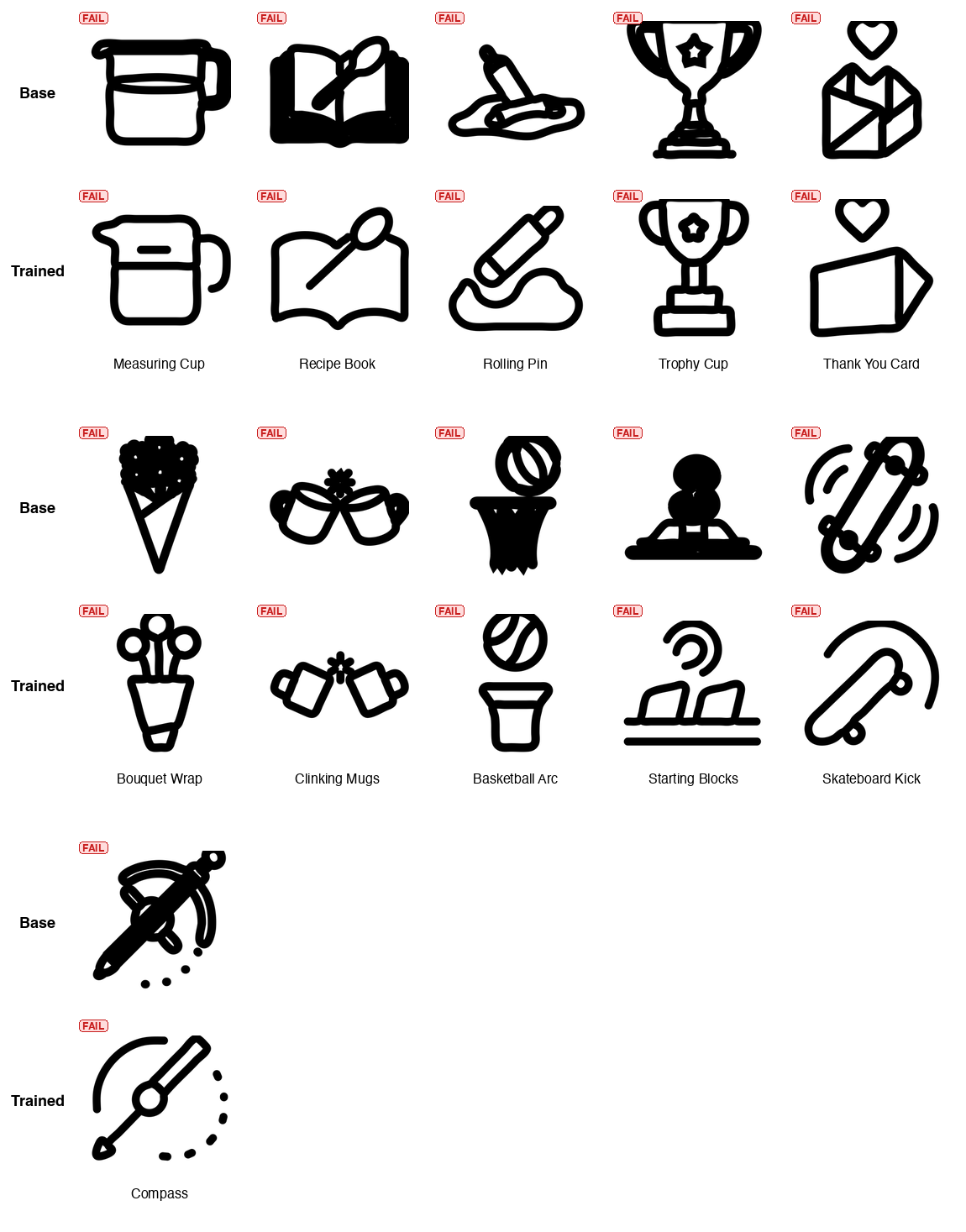}
\caption{Concepts where both the base model (top of each pair) and the
trained model (bottom) fail. Failures are either local junction merges on moderately
complex objects (measuring cup, trophy cup, thank you card) or complexity-driven: the
concept demands more visual structure than the model can represent legibly (basketball
arc, skateboard kick, compass).}
\label{fig:appendix-both-fail}
\end{figure*}

\paragraph{Both models fail.}
Figure~\ref{fig:appendix-both-fail} shows $16$ concepts where both models fail. The
failure modes split into two categories. Some concepts involve local junction merges: the
measuring cup handle meets the body, the rolling pin edges crowd against the dough, and
the trophy cup stem fuses with the base. The trained model draws these more cleanly than
the base but not enough to pass the VLM judge. Other concepts are inherently too complex:
the basketball arc, skateboard kick, and starting blocks require multiple overlapping
parts in action poses, producing too many strokes for the model to separate legibly. The
trained model simplifies these but often at the cost of recognisability, drawing shapes
that no longer convey the intended concept.

\end{document}